\documentclass[journal]{IEEEtran}
\ifCLASSINFOpdf
\else
\fi
  \usepackage{graphicx}
  \usepackage{amsmath,amssymb}
  \usepackage{color}

 \newcommand{\R}{\mathbb{R}}
  \newcommand{\dn}{\mathbf{d}}
 
 \newcommand{\diag}{\mathrm{diag}}

\newcommand{\dint} {\displaystyle\int}
\newcommand{\dsum} {\displaystyle\sum}

  \newcommand{\zero}{\mathbf{0}}
\newtheorem{theorem}{Theorem}

\begin{document}
%
\title{Homogeneous Proportional-Integral-Derivative Controller in Mobile Robotic Manipulators}
%
%
%

\author{
        Luis~Luna,~Isaac~Chairez,~\IEEEmembership{Member,~IEEE} and Andrey Polyakov
\thanks{Luis Luna and Isaac Chairez are with the Tecnologico de Monterrey, Institute of Advanced Materials for Sustinable Manufacturing, Zapopan, Jalisco, Mexico,  e-mail: isaac.chairez@tec,mx. }
\thanks{Andrey Polyakov is with Inria Centere of the University of Lille, CNRS CRIStAL, FR-59000, Lille, France. e-mail:}
\thanks{Manuscript received April 19, 2005; revised August 26, 2025.}}

%
%

\markboth{IEEE Transactions on System, Man, and Cybernetics,~Vol.~14, No.~8, August~2025}%
{Shell \MakeLowercase{\textit{et al.}}: Bare Demo of IEEEtran.cls for IEEE Journals}
%



\maketitle

\begin{abstract}
Mobile robotic manipulators (MRMs), which integrate mobility and manipulation capabilities, present significant control challenges due to their nonlinear dynamics, underactuation, and coupling between the base and manipulator subsystems. This paper proposes a novel homogeneous Proportional-Integral-Derivative (hPID) control strategy tailored for MRMs to achieve robust and coordinated motion control. Unlike classical PID controllers, the hPID controller leverages the mathematical framework of homogeneous control theory to systematically enhance the stability and convergence properties of the closed-loop system, even in the presence of dynamic uncertainties and external disturbances involved into a system in a homogeneous way. A homogeneous PID structure is designed, ensuring improved convergence of tracking errors through a graded homogeneity approach that generalizes traditional PID gains to nonlinear, state-dependent functions. Stability analysis is conducted using Lyapunov-based methods, demonstrating that the hPID controller guarantees global asymptotic stability and finite-time convergence under mild assumptions. Experimental results on a representative MRM model validate the effectiveness of the hPID controller in achieving high-precision trajectory tracking for both the mobile base and manipulator arm, outperforming conventional linear PID controllers in terms of response time, steady-state error, and robustness to model uncertainties. This research contributes a scalable and analytically grounded control framework for enhancing the autonomy and reliability of next-generation mobile manipulation systems in structured and unstructured environments.
\end{abstract}


%
\IEEEpeerreviewmaketitle

\section{Introduction}

The integration of mobility and manipulation in robotic systems has become increasingly significant in a broad spectrum of modern industrial and service applications. Mobile robotic manipulators (MRMs), which combine the locomotion capabilities of mobile platforms with the dexterity of robotic arms, provide a unique solution for tasks that require both navigation and complex interaction with dynamic environments. From warehouse automation and space exploration to surgical robotics and agricultural automation, MRMs are at the forefront of robotics research and development. However, controlling such systems presents a complex challenge due to the hybrid nature of their dynamics and the need for coordinated control between the base and the manipulator.

Traditional control strategies for robotic systems have often relied on the Proportional-Integral-Derivative (PID) controller, a feedback mechanism that has demonstrated effectiveness over the last century due to its simplicity, ease of implementation, and robust performance in the presence of modeling uncertainties. PID controllers have been widely used in both linear and nonlinear control systems, including robotic manipulators and mobile platforms. However, when applied to MRMs, standard PID control may exhibit limitations, especially in handling the highly coupled, nonlinear, and non-holonomic dynamics that characterize such systems.

Recent advances in control theory have sought to enhance the performance of PID-based controllers in complex robotic applications. One promising avenue is the development of homogeneous PID controllers (see, e.g., \cite{Nakamura2013:SICE}, \cite{Polyakov2025:Book_Vol_I}, \cite{Fukui2021:SICE}) that is a class of control strategies derived from the concept of homogeneity in nonlinear systems. Homogeneous control techniques exploit scale-invariance properties and allow faster (finite/fixed-time) transients, better robustness and smaller overshoots comparing with linear control strategies (see, e.g., \cite[Chapter 1]{Polyakov2025:Book_Vol_I} for more details). The homogeneity can be highly advantageous in achieving precise and robust control in systems subject to disturbances, parametric uncertainties, and external perturbations. The homogeneous PID controller represents a natural extension of the classical PID framework, but with nonlinear components designed to preserve homogeneity and ensure desirable stability characteristics.

Homogeneity is a dilation symmetry (scaling invariance) known since the works of Leonhard Euler published in the 18th century. Generalized homogeneity known today as the weighted homogeneity has been introduced in the paper \cite{Zubov1958:IVM}. 
In  control systems  theory the weighted homogeneity is utilized for
stability analysis and Lyapunov function design \cite{Zubov1958:IVM}, \cite{Rosier1992:SCL}, \cite{BhatBernstein2005:MCSS}, \cite{Aleksandrov_etal2012:SCL};
stabilizability/controllability/observability  analysis
\cite{Hermes1986:SIAM_JCO}, \cite{Kawski1990:CTAT},  
\cite{SepulchreAeyels1996:MCSS}, \cite{SepulchreAeyels1996:SIAM_JCO};
 controllers/observers design 
 \cite{Andreini_etal1988:SCL}, \cite{CoronPraly1991:SCL},  \cite{Grune2000:SIAM_JCO}, 
 \cite{Perruquetti_etal2008:TAC}, \cite{Andrieu_etal2008:SIAM_JCO}, \cite{Polyakov_etal2015:Aut}, \cite{Lopez-Ramirez_etal2018:Aut};
robustness analysis \cite{Ryan1995:SCL}, \cite{Hong2001:Aut}, \cite{Andrieu_etal2008:SIAM_JCO}; 
sliding mode control and estimation  \cite{Levant2005:Aut}, 
\cite{Orlov2005:SIAM_JCO}, \cite{MorenoOsorio2012:IEEE_TAC}, \cite{Bernuau_etal2014:JFI}, \cite{CruzMoreno2017:Aut};
optimal and model predictive control  \cite{ZelikinBorisov1994:Book},  \cite{Coron_etal2019:SIAM_JCO}, \cite[Chapter 13]{Polyakov_etal2025:Book_Vol_I};
digitization of the control systems \cite{LivneLevant2014:Aut},
\cite{Koch_etal2019:TAC}, \cite{Polyakov_etal2023:Aut}.
All mentioned  and may other problems of homogeneous control theory are studied in the monograph \cite{Polyakov2025:Book_Vol_I} using the so-called  linear dilations. The theory of homogeneous control systems is well-developed in the literature. However, just a few practical applications of homogeneous control algorithms have been reported \cite{Nakamura2013:SICE}, \cite{Wang_etal2020:ICRA}, \cite{Cruz-Ortiz_etal2021:TIE}, \cite{Zhou_etal2024:TCST}. All of them demonstrate a potential usefulness of homogeneous PD/PID controllers with negative degrees. One of aims of this paper is to investigate if homogeneous PID with positive degree may guarantee a better regulation quality comparing with the conventional linear PID. For this purpose we develop a special procedure for transformation of linear PID to a homogeneous one.
The procedure refines the ideas proposed in \cite{Wang_2020:RNC}, \cite{Cruz-Ortiz_etal2021:TIE} and essentially extends the class of admissible homogeneous PID regulators.

The application of homogeneous PID control to mobile robotic manipulators is still a relatively unexplored domain. Most existing approaches either focus on the mobile platform or the manipulator arm in isolation, often neglecting the intricate coupling between the two subsystems. Moreover, many conventional control schemes do not explicitly address the inherent nonholonomic constraints of mobile platforms, such as differential drive or car-like kinematics, which complicate the trajectory tracking and posture regulation tasks. The challenge lies in developing a unified control scheme that not only handles the coupled kinematics and dynamics of the MRM but also ensures convergence and robustness under realistic operational conditions.

In this context, the present paper proposes a novel homogeneous PID controller for mobile robotic manipulators, designed to address the limitations of conventional PID controllers and provide enhanced performance in nonlinear, uncertain, and coupled robotic systems. The proposed controller leverages the theory of finite-time stable systems and homogeneous feedback design to achieve accurate and robust trajectory tracking for both the mobile base and the manipulator arm. By incorporating nonlinear proportional, integral, and derivative actions, the controller ensures better disturbance rejection and faster convergence compared to classical linear PID schemes.

This work is motivated by the growing need for advanced control techniques that can be deployed in real-world applications where MRMs operate in unstructured environments, interact with uncertain objects, and perform complex tasks autonomously. The homogeneous PID controller introduced here is constructed using a systematic approach that accounts for the hierarchical structure of the MRM system, its kinematic and dynamic constraints, and the interdependence of motion between the base and the arm. A key feature of the controller is its ability to guarantee convergence in finite time, a property that enhances both responsiveness and safety in time-critical applications such as human-robot interaction or cooperative manipulation.

To validate the effectiveness of the proposed control strategy, a detailed theoretical analysis is conducted, including proofs of stability and convergence tailored to homogeneous systems. Additionally, the controller is implemented and tested in a hardware-in-the-loop set of experiments involving a robotic manipulator based on a mobile base and a six-degree-of-freedom robotic arm. 

The crucial contributions of this study can be summarized as follows:
\begin{itemize}
    \item A novel homogeneous PID controller, being different from the one studied in the literature \cite{Nakamura2013:SICE}, \cite{Fukui2021:SICE}, is introduced. The controller is defined by means of the homogeneous norm in $\R^2$.    
    \item The main theoretical contribution of the paper is the proof  that a homogeneous PID regulator can be obtained from any linear PID controller by means of a proper tuning of the homogeneity degree. Moreover, such a transformation ("upgrade") is successful independently of a selection of the homogeneous norm.
    \item  Experimental confirmation of hPID improved tracking trajectory quality compared to regular linear PID on a mobile robotic manipulator
\end{itemize}

{This manuscript is organized as follows: 
First, some preliminaries about homogeneity are given. Next, a homogeneous PID controller is introduced and studied. After that the hPID is applied for control of robotic manipulator. The experimental results comparing linear PID with homogeneous PID (obtained from this linear PID) are presented.  Finally some concluding remarks are presented.}

\section{Mathematical preliminaries on homogeneity}

\subsection{Linear Dilation}
Let us recall  that \textit{a family of  operators} $\dn(s):\R^n\mapsto \R^n$ with $s\in \R$ is  a one-parameter \textit{group} if
\begin{itemize}
\item $\dn(0)x\!=\!x$ for all $x\!\in\!\R^n$;
\item $\dn(s) \dn(t) x\!=\!\dn(s+t)x$ for all $x\!\in\!\R^n$ and all $s,t\!\in\!\R$.
\end{itemize}
A \textit{group} $\dn$ is 
\begin{itemize}
\item \textit{continuous} if the mapping $s\mapsto \dn(s)x$ is continuous for any  $x\!\in\! \R^n$;
\item \textit{linear} if $\dn(s)\in \R^{n\times n}$ for all $s\in \R$;
\item a \textit{dilation} \cite{Kawski1991:ACDS} in $\R^n$ if $\liminf\limits_{s\to +\infty}\|\dn(s)x\|=+\infty$ and $\limsup\limits_{s\to -\infty}\|\dn(s)x\|=0$ for all $x\neq \zero$.
\end{itemize}
It is well known (see, e.g., \cite[Chapter1, Theorem 1.2]{Pazy1983:Book}) that any linear continuous group in $\R^n$ admits the representation 
	$\dn(s)=e^{sG_{\dn}}=\sum_{j=1}^{\infty}\tfrac{s^jG_{\dn}^j}{j!}$,
where $G_{\dn}\in \R^{n\times n}$ is a \textit{generator of the group} $\dn$. Such a  continuous linear group  is a dilation in $\R^n$ if and only if $G_{\dn}$ is an anti-Hurwitz matrix \cite[Theorem 5.1]{Polyakov2025:Book_Vol_I}.  In this paper, we deal only with continuous linear dilation having a diagonal generator
$G_{\dn}=\diag(r_1,r_2,...,r_n)$, where 
the positive parameters $r_1,...,r_n>0$ define the so-called dilation weights. Such a dilation 
 is known in the literature as \textit{weighted dilation} (see, e.g., \cite{Zubov1958:IVM}, \cite{Rosier1992:SCL}, \cite{Grune2000:SIAM_JCO})
and has the form
\begin{equation}
\dn(s)=\diag(e^{r_1s},e^{r_2s},...,e^{r_ns}).
\end{equation}
A \textit{dilation} $\dn$ in $\R^n$ is
\begin{itemize} 
\item \textit{monotone} with respect to the norm $\|\cdot\|$ if the function $s\mapsto \|\dn(s)x\|$ is strictly increasing for all  $x\neq \zero$;
\item 	\textit{strictly monotone} with respect to the norm $\|\cdot\|$  if  $\exists \beta\!>\!0$ such that $\|\dn(s)x\|\!\leq\! e^{\beta s}\|x\|$ for all  $ s\!\leq\! 0$ and all $x\in \R^n$.
\end{itemize}
One can be shown \cite[Proposition 5.3]{Polyakov2025:Book_Vol_I} that  a linear continuous dilation in $\R^n$ is strictly monotone with respect to the weighted Euclidean norm $\|x\|=\sqrt{x^{\top} Px}$ with $0\prec P\in \R^{n\times n}$ if and only if 
$	PG_{\dn}+G_{\dn}^{\top}P\succ 0,  P\succ 0.$

\subsection{Homogeneous norm}

A continuous positive definite function $\|\cdot\|_{\dn}: \R^n\mapsto \R$ satisfying $\|\pm\dn(s)x\|=e^{s}\|x\|_{\dn}, \forall x\in \R^n$ is usually called  a $\dn$-\textit{homogeneous norm}
(or ``dilated norm'')  \cite{Kawski1991:ACDS}, \cite{Grune2000:SIAM_JCO}, \cite{Hong2001:Aut}, \cite{Andrieu_etal2008:SIAM_JCO}. 
For the weighted dilation, a $\dn$-homogeneous norm can be defined, for example, as 
\begin{equation}
    \|x\|_{\dn}=\sum_{i=1}^{n}\gamma_i |x_i|^{\frac{1}{r_i}},
\end{equation}
where $x=(x_1,...,x_n)^{\top}\in \R^n$ and $\gamma_1,...,\gamma_n>0$. For $\gamma_1=...=\gamma_n=1$ such a $\dn$-homogeneous norm is considered, for example, in \cite{Kawski1991:ACDS}.

The $\dn$-homogeneous norm $\|\cdot\|_{\dn}$ does not satisfy the triangle inequality in the general case,  so, formally, this is not even a semi-norm. However, we would follow the tradition accepted in systems sciences and call the function $\|\cdot\|_{\dn}$ satisfying the above property by a ``norm'', but, basically, we use this name for the  canonical homogeneous norm (see below) being a norm (in the classical sense) for the vector space $\R^n_{\dn}$ homeomorphic to $\R^n$ \cite[Theorem 5.3]{Polyakov2025:Book_Vol_I}. 

	Let a linear continuous dilation $\dn$ in $\R^n$ be monotone with respect to a norm $\|\cdot\|$.
    Following \cite[Chapter 5]{Polyakov2025:Book_Vol_I},
	a function $\|\cdot\|_{\dn} : \R^n \mapsto [0,+\infty)$ defined by 
	\begin{equation}\label{eq:hom_norm_Rn}
		\|x\|_{\dn}\!=\!\left\{
        \begin{array}{ccc}
        0 & \text{ if } & x=\zero;\\
        \mathop{\mathrm{argmin}}\limits_{\lambda>0} (\|\dn(-\ln \lambda)x\|-1)^2 & \text{ if } & x\neq \zero.
        \end{array}
        \right.
	\end{equation}
	is said to be a canonical $\dn$-homogeneous norm in  $\R^n$. Canonical homogeneous norm have may important properties useful for control systems design
    (see, e.g., \cite[Chapters 5 \& 9]{Polyakov2025:Book_Vol_I}). For example, if the canonical homogeneous norm 
    $\|x\|_{\dn}$ is induced by the weighted Euclidean norm $\|x\|=\sqrt{x^{\top}Px}$ with $0\prec P=P^{\top}\in \R^{n\times n} :PG_{\dn}+G_{\dn}^{\top}P\succ 0$ then 
    \begin{equation}\label{eq:der_hnorm}
    \tfrac{\partial \|x\|_{\dn}}{\partial x}=\|x\|_{\dn}\tfrac{x^{\top}\dn^{\top}(-\ln \|x\|_{\dn})P\dn(-\ln \|x\|_{\dn})}{x^{\top}\dn^{\top}(-\ln \|x\|_{\dn})PG_{\dn }\dn(-\ln \|x\|_{\dn})x}, \quad \forall x\neq \zero
    \end{equation}
For the standard dilation $\dn(s)\!=\!e^{s}I_n$, we have $\|x\|_{\dn}\!=\!\|x\|$. 
In other cases, the canonical homogeneous norm $\|x\|_{\dn}$ with $x\neq \zero$ is implicitly defined by the nonlinear algebraic equation $\|\dn(-\ln \lambda)x\|=1$ with respect $\lambda>0$, which always has a unique solution  due to monotonicity of the dilation. Notice that 
all $\dn$-homogeneous norms are equivalent \cite[Corollary 5.4]{Polyakov2025:Book_Vol_I}, i.e., for any pair of $\dn$-homogeneous norms $\|\cdot\|_{\dn,1}$ and $\|\cdot\|_{\dn,2}$, there exist $0<k_1\leq k_2<+\infty$ such that 
$
k_1\|x\|_{\dn,1}\leq \|x\|_{\dn,2}\leq k_2\|x\|_{\dn,1},  \forall x\in \R^n.
$  

\subsection{Homogeneous systems}
The  dilation symmetry of functions and vector fields in $\R^n$ is introduced by the following relations known as conditions of generalized  homogeneity \cite{Zubov1958:IVM}, \cite{Kawski1991:ACDS}, \cite{Rosier1992:SCL}, \cite{BhatBernstein2005:MCSS}, \cite{Polyakov2025:Book_Vol_I}:
	\begin{itemize}	
        \item A function $h: \R^n\to \R$ is $\dn$-homogeneous of degree $\nu\!\in\! \R$ if 
		\[
		h(\dn(s)x)=e^{\nu s}h(x), \quad \forall x\in \R^n, \quad \forall s\in \R.
		\]
	\item A vector field $g:\R^n \to \R^n$ is  $\dn$-homogeneous of 
		degree $\mu\in \R$  if 
		\begin{equation}\label{eq:homogeneous_operator_Rn}
			g(\dn(s)x)=e^{\mu s}\dn(s)g(x), \quad  \quad  \forall s\in\R, \quad \forall x\in \R^n.
		\end{equation}
        \item An ODE 
        \begin{equation}\label{eq:ODE}
        \dot x=g(x), \quad t>0, \quad x(t)\in \R^n 
        \end{equation}is said to be $\dn$-homogeneous of degree $\mu$ if the corresponding vector field $g$ is $\dn$-homogeneous of degree $\mu$.
        \end{itemize}
For simplicity, we consider only ODEs with continuous vector fields, i.e., $g\in C(\R^n,\R^n)$. The homogeneity of the system implies equivalence of its local and global properties \cite{Zubov1958:IVM}. For example, the local stability
of the $\dn$-homogeneous ODE \eqref{eq:ODE} is equivalent to its global  stability \footnote{The system \eqref{eq:ODE} is said to be (see, e.g., \cite{EfimovPolyakov2021:Book})
\begin{itemize} 
\item globally Lyapunov stable if $\exists \alpha\in\mathcal{K}_{\infty}$ such that
$\|x(t,x_0)\|\leq \alpha(\|x_0\|)$ for all $t\geq 0$ and all $x_0\in \R^n$;
\item globally asymptotically stable if it is Lyapunov stable and $x(t,x_0)\to 0$ as $t\to +\infty$;
\item globally exponentially stable if exist $C\geq 1$ and $r>0$ such that
$\|x(t,x_0)\|\leq Ce^{-r t}\|x_0\|$ for all $t\geq 0$ and all $x_0\in \R^n$;
\item globally finite-time stable if it is Lyapunov stable and for any $x_0\in \R^n$ there exists  $T(x_0)\geq 0$ such that $x(t,x_0)=\zero$ for all $t\geq T(x_0)$;
\item globally nearly fixed-time stable if it is Lyapunov stable and for any $r>0$ there exists $T_r>0$ such that $\|x(t,x_0)\|\leq r$ for all $t\geq T_r$ and all $x_0\in \R^n$.
\end{itemize}} due to the dilation symmetry of solutions:
\[
x(t,\dn(s)x_0)=\dn(s)x(e^{\mu s}t,x_0), \; \forall t\geq 0, \; \forall x_0\in \R^n, \; \forall s\in \R,
\]
where $x(\cdot,x_0)$ denotes a solution of \eqref{eq:ODE} with the initial condition 
$x(0)=x_0$, but $x(\cdot,\dn(s)x_0)$ denotes a solution of \eqref{eq:ODE} with the scaled initial condition 
$x(0)=\dn(s)x_0$.

The homogeneity degree of the system  specifies its convergence rate, namely, if the $\dn$-homogeneous ODE \eqref{eq:ODE} is asymptotically stable then \cite{Zubov1964:Book}, \cite{Nakamura_etal2002:SICE}, \cite[Theorem 7.5]{Polyakov2025:Book_Vol_I}
\begin{itemize}
\item for $\mu<0$ it is globally finite-time stable;
\item for $\mu=0$ it is globally exponentially stable;\vspace{1mm}
\item for $\mu>0$ it is globally nearly fixed-time stable.
\end{itemize}
In the view of above result, homogeneous systems may have faster convergence than linear system in the case of non-zero homogeneity degree. In addition to faster convergence, the homogeneous systems may demonstrate also a better robustness and smaller overshoots than linear systems (see, \cite[Chapter 1]{Polyakov2025:Book_Vol_I}  for more details). 

Homogeneous systems are robust (Input-to-State Stable\footnote{The system \eqref{eq:sys_pert} is said to be Input-to-State Stable (ISS) \cite{Sontag1989:TAC} if there exists 
$\gamma\in \mathcal{K}$ and $\beta\in \mathcal{KL}$ such that 
\[
\|x(t,x_0)\|\leq \beta(\|x_0\|,t)+\gamma\left(\|q\|_{L^{\infty}_{(0,t)}}\right)
\]
for all $x_0\in \R^n$ and solutions $x(t,x_0)$ of the system \eqref{eq:sys_pert}.
}) with respect homogeneously involved perturbations \cite{Ryan1995:SCL}, \cite{Andrieu_etal2008:SIAM_JCO}. Namely \cite[Theorem 7.13]{Polyakov2025:Book_Vol_I}, if $f: \R^n\times \R^k\to \R^n$ is a continuous vector field and there exist two linear dilations 
$\dn$ in $\R^n$ and $\hat \dn$ in $\R^k$ such that
\[
f(\dn(s)x,\hat \dn(s)q)=e^{\mu s}\dn(s)f(x,q), \quad \forall x\in \R^n,\forall q\in \R^k
\]
then the global asymptotic stability of the unpertubed system
$\dot x=f(x,\zero)$ implies the input-to-state stability of the perturbed system
\begin{equation}\label{eq:sys_pert}
\dot x\!=\!f(x, q),\;\; t>0,\;\; x(0)\!=\!x_0\!\in\! \R^n, \;\; q\!\in\! L^{\infty}(\R,\R^k).
\end{equation}
The latter property of homogeneous systems means that a robust homogeneous control system can be designed omitting perturbations provided that they are incorporated in the system is homogeneous manner. 
\section{Homogeneous PID control}
\subsection{Design of hPID}
Considering a control system with a single input $u(t)$ and a single output $y(t)$, the following feedback law
\begin{equation}\label{eq:PID}
u_{\rm PID}\!=\!K_{\rm p} \epsilon\!+\!K_{\rm d} \dot \epsilon\!+\!K_{\rm i}\int \!\!\epsilon(\tau) d\tau, \quad\quad  K_{\rm p},K_{\rm d},K_{\rm i}\!\in\! \R,\;\;
\end{equation}
is known as a Proportional-Integral-Derivative (PID) regulator (see, e.g., \cite{AstromHagglund1995:Book}), where
\[
\epsilon(t)=y(t)-y^*(t)
\]
is a regulation error and $y^*(t)$ is a desired value of the output of the control, which may, probably, be time-varying. Inspired by \cite{Zhou_etal2023:RNC}, \cite[Chapter 11]{Polyakov2025:Book_Vol_I}, let us consider the following feedback law
\begin{equation}\label{eq:hPID}
 u_{\rm hPID}= K_{\rm p} \|\xi\|_{\dn}^{2\mu}\epsilon+K_{\rm d} \|\xi\|_{\dn}^{\mu}\dot \epsilon+K_{\rm i}\int \|\xi\|_{\dn}^{3\mu} \epsilon \, d\tau,
\end{equation}
where $\mu\in(-0.5,0.5)$, $\dn(s)=\diag (e^{(1-\mu)s},e^s)$,
$$\xi=\left(
\begin{array}{c}
\epsilon\\
\dot \epsilon
\end{array}
 \right),
 $$
and $\|\cdot\|_{\dn}$ is a $\dn$-homogeneous norm. Being essentially different from the fractional polynomial feedback law considered in \cite{Nakamura2013:SICE}, \cite{Fukui2021:SICE}, the feedback law 
\eqref{eq:hPID} can also be called as \textit{homogeneous PID regulator}.  Assuming that the dynamics of the error equation is given by the double integrator
\begin{equation}\label{eq:model}
\ddot \epsilon=u+p,  \quad p=\text{const}
\end{equation}
one can be easily shown that the closed-loop system is homogeneous of the degree $\mu$. Denoting $x=(x_1,x_2,x_3)^{\top}$, 
$$
x_1=\epsilon,\quad  x_2=\dot \epsilon,\quad  x_3=p+K_i\int \|\xi\|_{\dn}^{3\mu} \epsilon \,d\tau,
$$ for the closed-loop system \eqref{eq:model}, \eqref{eq:hPID} we have
\begin{equation}\label{eq:sys_with_hPID}
\dot x=g_{\mu}(x):=\left(
\begin{smallmatrix}
 x_2\\
 K_{\rm p} \|\xi\|_{\dn}^{2\mu}x_1+K_{\rm d} \|\xi\|_{\dn}^{\mu}x_2+x_3\\
 K_{\rm i} \|\xi\|_{\dn}^{3\mu}x_1
\end{smallmatrix}
\right), \quad \xi=\left(
\begin{smallmatrix}
x_1\\
x_2
\end{smallmatrix}
 \right)
\end{equation}
Since, $\|\dn(s)\xi\|_{\dn}=e^s \|\xi\|_{\dn}$ then
the vector field $f$ is $\tilde \dn$-homogeneous with respect to the dilation 
$$
\tilde {\dn}(s)=\diag(e^{(1-\mu)s}, e^{s}, e^{(1+\mu)s}),$$
namely,
\[
\begin{split}
g_{\mu}(\tilde{\dn}(s)x)\!=&\!\left(\!
\begin{smallmatrix}
 e^sx_2\\
 K_{\rm p} \|\dn(s)\xi\|_{\dn}^{2\mu}e^{(1-\mu)s}x_1+K_{\rm d} \|\dn(s)\xi\|_{\dn}^{\mu}e^{s}x_2+e^{(1+\mu)s }x_3\\
 K_{\rm i} \|\dn(s)\xi\|_{\dn}^{3\mu}e^{(1-\mu)s}x_1
\end{smallmatrix}
\!\right)\\
=& \left(
\begin{smallmatrix}
 e^sx_2\\
 e^{(1+\mu)s} K_{\rm p} \|\xi\|_{\dn}^{2\mu}x_1+e^{(1+\mu)s}K_{\rm d} \|\xi\|_{\dn}^{\mu}x_2+e^{(1+\mu)s }x_3\\
 e^{(1+2\mu)s}K_{\rm i} \|\xi\|_{\dn}^{3\mu}x_1
\end{smallmatrix}
\right)\\
=&e^{\mu s}\tilde{\dn}(s)g(x)
\end{split}
\]
for all $s\in \R$ and all $x\in \R^3$. Notice that $\tilde \dn(s)=e^{sG_{\tilde \dn}}$ with 
\begin{equation}
G_{\tilde \dn}=I_3+\mu\,\diag(-1,0,1).
\end{equation}

\subsection{Stability of hPID}
The hPID of the form \eqref{eq:hPID} was studied in \cite{Zhou_etal2023:RNC} in the case of the canonical homogeneous norm $\|\cdot\|_{\dn}$ induced by a weighted Euclidean norm. In this paper we show that any homogeneous $\|\cdot\|_{\dn}$ can be utilized in order to design a stable hPID \eqref{eq:hPID}.

For $\mu=0$ the hPID \eqref{eq:hPID} becomes the conventional (liner) PID \eqref{eq:PID}. By continuity, we may expect that if the linear PID asymptotically stabilizes the system \eqref{eq:model} then the homogeneous PID also does, at least, for $\mu$ being close to $0$. Let us prove this claim rigorously. 
\begin{theorem}\itshape
Let a linear dilation $\dn$ in $\R^2$ be defined by the formula $\dn(s)=\diag (e^{(1-\mu)s},e^s)$, $s\in \R$.
Let $\|\cdot\|_{\dn}$ be an arbitrary $\dn$-homogeneous norm in $\R^2$.
If  the system \eqref{eq:model} with the linear PID \eqref{eq:PID} is globally asymptotically stable, then the system \eqref{eq:model} with the homogeneous PID \eqref{eq:hPID} is globally asymptotically stable, at least, for $\mu$ being close to zero.
\end{theorem}
\textbf{Proof.} By assumption the system
\[
\dot x=g_0(x)=\underbrace{\left(\begin{array}{ccc}
0 & 1 & 0\\
K_{\rm p} & K_{\rm  d} & 1\\
K_{\rm i} & 0 & 0 
\end{array}\right)}_{=A}x
\]
is globally asymptotically stable. In this case, the Lyapunov inequality \cite{Lyapunov1892:Book} 
\[
PA+A^{\top} P\prec 0, \quad P\succ0
\]
is feasible with respect to $P=P^{\top}\in \R^{3\times 3}$. Since 
\[
PG_{\tilde \dn}+G_{\tilde \dn}^{\top}P=2P+\mu (P\diag(-1,0,1)+\diag(-1,0,1)P)
\]
then for $\mu$ being sufficiently close to $0$ we have
\[
PG_{\tilde \dn}+G_{\tilde \dn}^{\top}P\succ 0.
\]
Let us show that the canonical homogeneous norm $\|x\|_{\tilde\dn}$ induced by the weighted Euclidean norm $\|x\|=\sqrt{x^{\top}Px}$ is the Lyapunov function for the closed-loop system \eqref{eq:model} with the homogeneous PID \eqref{eq:hPID},
which has the form \eqref{eq:sys_with_hPID}. Indeed, using the formula 
\eqref{eq:der_hnorm} for $x\neq \zero$, we derive
\[
\begin{split}
\tfrac{d}{dt}\|x\|_{\tilde \dn}=&\tfrac{\partial \|x\|_{\tilde \dn}}{\partial x} g_{\mu}(x) \\
=& \|x\|_{\tilde \dn}\tfrac{x^{\top}\tilde \dn^{\top}(-\ln \|x\|_{\tilde \dn})P\tilde \dn(-\ln \|x\|_{\tilde \dn})g_{\mu}(x)}{x^{\top}\tilde \dn^{\top}(-\ln \|x\|_{\tilde \dn})PG_{\tilde \dn }\tilde \dn(-\ln \|x\|_{\tilde \dn})x}\\
=&\|x\|_{\tilde \dn}^{1+\mu}\tfrac{x^{\top}\tilde \dn^{\top}(-\ln \|x\|_{\tilde \dn})Pg_{\mu}(\tilde \dn(-\ln \|x\|_{\tilde \dn})x)}{x^{\top}\tilde \dn^{\top}(-\ln \|x\|_{\tilde \dn})PG_{\tilde \dn }\tilde \dn(-\ln \|x\|_{\tilde \dn})x}\\
=&\|x\|_{\tilde \dn}^{1+\mu}\tfrac{x^{\top}\tilde \dn^{\top}(-\ln \|x\|_{\tilde \dn})PA\dn(-\ln \|x\|_{\tilde \dn})x}{x^{\top}\tilde \dn^{\top}(-\ln \|x\|_{\tilde \dn})PG_{\tilde \dn }\tilde \dn(-\ln \|x\|_{\tilde \dn})x}\\
&+\|x\|_{\tilde \dn}^{1+\mu}\tfrac{x^{\top}\tilde \dn^{\top}(-\ln \|x\|_{\tilde \dn})P\Delta g(\tilde \dn(-\ln \|x\|_{\tilde \dn})x)}{x^{\top}\tilde \dn^{\top}(-\ln \|x\|_{\tilde \dn})PG_{\tilde \dn }\tilde \dn(-\ln \|x\|_{\tilde \dn})x},
\end{split}
\]
where 
\[
\Delta g(z):=g_{\mu}(z)-g_0(z),\quad z\in \R^3.
\]
Denoting 
\[
\beta=\lambda_{\min} (P^{\frac 1 2}G_{\tilde \dn}P^{-\frac 1 2}+P^{-\frac 1 2}G_{\tilde \dn}^{\top}P^{\frac 1 2})>0
\]
we derive
\[
x^{\top}\tilde\dn^{\top}(-\ln \|x\|_{\tilde \dn})PG_{\tilde \dn }\tilde \dn(-\ln \|x\|_{\tilde \dn})x
\]
\[
=\tfrac{x^{\top}\tilde \dn^{\top}(-\ln \|x\|_{\tilde \dn})P^{\frac 1 2}\left(P^{\frac 1 2}G_{\tilde \dn}P^{-\frac 1 2}+P^{-\frac 1 2}G_{\tilde \dn}^{\top}P^{\frac 1 2}\right)P^{\frac 1 2}
\tilde \dn(-\ln \|x\|_{\tilde \dn})x}{2}
\]
\[
 \ge \frac{ \beta x^{\top}\tilde \dn^{\top}(-\ln \|x\|_{\tilde \dn})P\tilde \dn(-\ln \|x\|_{\tilde \dn})x}{2}=-\frac{\beta}{2},
\]
where the identity $x^{\top}\tilde \dn^{\top}(-\ln \|x\|_{\tilde \dn})P\tilde \dn(-\ln \|x\|_{\tilde \dn})x=1$
(see, the definition of the canonical homogeneous norm) is utilized on the last step.
Similarly, for
\[
\gamma=-\lambda_{\max} (P^{\frac 1 2}AP^{-\frac 1 2}+P^{-\frac 1 2}A^{\top}P^{\frac 1 2})>0
\]
we have 
\[
x^{\top}\tilde \dn^{\top}(-\ln \|x\|_{\tilde \dn})PA\tilde \dn(-\ln \|x\|_{\tilde \dn})\tilde x
\]
\[
=\tfrac{x^{\top}\tilde \dn^{\top}(-\ln \|x\|_{\tilde \dn})P^{\frac 1 2}\left(P^{\frac 1 2}AP^{-\frac 1 2}+P^{-\frac 1 2}A^{\top}P^{\frac 1 2}\right)P^{\frac 1 2}
\dn(-\ln \|x\|_{\tilde \dn})x}{2}
\]
\[
\le\frac{\lambda_{\max} (P^{\frac 1 2}AP^{-\frac 1 2}+P^{-\frac 1 2}A^{\top}P^{\frac 1 2})}{2}=-\frac{\gamma}{2}.
\]
Therefore, taking into account
\[
x^{\top}\tilde \dn^{\top}(-\ln \|x\|_{\tilde \dn})P\Delta g(\tilde \dn(-\ln \|x\|_{\tilde \dn})x)
\]
\[
\le\underbrace{\|\tilde \dn(-\ln \|x\|_{\tilde \dn})x\|}_{=1}\cdot \|\Delta g(\tilde \dn(-\ln \|x\|_{\tilde \dn})x)\|
\]
we derive
\[
\tfrac{d}{dt}\|x\|_{\tilde \dn}\leq \frac{-\gamma + 2\|\Delta g(\tilde \dn(-\ln \|x\|_{\tilde \dn})x)\|}{\beta}\|x\|_{\tilde \dn}^{1+\mu}.
\]
Since $\|\tilde \dn(-\ln \|x\|_{\tilde \dn})x\|=1$ then, to complete the proof, we just need to show that 
\begin{equation}\label{eq:limit_delta_g}
\sup_{z^{\top}Pz=1} \Delta g(z) \to 0 \text{ as } \mu \to 0.
\end{equation}
Notice that 
\[
 \Delta g(z)=
 \left(
\begin{smallmatrix}
 0\\
 K_{\rm p} \left(\left\|\tilde \xi \right\|_{\dn}^{2\mu}-1\right)z_1+K_{\rm d} \left(\left\|\tilde \xi \right\|_{\dn}^{\mu}-1\right)z_2\\
 K_{\rm i} \left(\left\|\tilde \xi\right\|_{\dn}^{3\mu}-1\right)z_1
\end{smallmatrix}
\right)
\]
\[
= \left(
\begin{smallmatrix}
 0\\
 K_{\rm p} \left(\left\|\tilde \xi \right\|_{\dn}^{1+\mu}-\left\|\tilde \xi \right\|_{\dn}^{1-\mu}\right)\frac{z_1}{\left\|\tilde \xi \right\|_{\dn}^{1-\mu}}+K_{\rm d} \left(\left\|\tilde \xi\right\|_{\dn}^{1+\mu}-\left\|\tilde \xi \right\|_{\dn}\right)\frac{z_2}{\left\|\tilde \xi \right\|_{\dn}}\\
K_{\rm i} \left(\left\|\tilde \xi \right\|_{\dn}^{1+2\mu}-\left\|\tilde \xi \right\|_{\dn}^{1-\mu}\right)\frac{z_1}{\left\|\tilde \xi \right\|_{\dn}^{1-\mu}}
\end{smallmatrix}
\right),
\]
where $\tilde \xi=\left(\begin{smallmatrix}
     z_1\\z_2
 \end{smallmatrix}\right)$. Since
 \[
 \|\dn(-\ln \|\tilde \xi\|_{\dn})\xi\|_{\dn}=\left\|
 \left(
 \begin{smallmatrix}
     \frac{z_1}{\left\|\tilde \xi \right\|_{\dn}^{1-\mu}}\\
     \frac{z_2}{\left\|\tilde \xi \right\|_{\dn}}
 \end{smallmatrix}
 \right)
 \right\|=1 \quad \text{ for all } \quad \tilde \xi \neq \zero,
 \]
 then the terms $\left|\frac{z_1}{\left\|\tilde \xi \right\|_{\dn}^{1-\mu}}\right|$ and $\left|\frac{z_2}{\left\|\tilde \xi \right\|_{\dn}}\right|$ are uniformly bounded.
 Taking into account that
 \[
\sup_{z^{\top}Pz=1} \left|\left\|\tilde \xi \right\|_{\dn}^{1+\mu}-\left\|\tilde \xi \right\|_{\dn}^{1-\mu}\right|\to 0 \text{ as } \mu \to 0
 \]
  \[
\sup_{z^{\top}Pz=1}  \left|\left\|\tilde \xi \right\|_{\dn}^{1+\mu}-\left\|\tilde \xi \right\|_{\dn}\right|\to 0 \text{ as } \mu \to 0
 \]
  \[
\sup_{z^{\top}Pz=1}  \left|\left\|\tilde \xi \right\|_{\dn}^{1+2\mu}-\left\|\tilde \xi \right\|_{\dn}^{1-\mu}
\right|\to 0 \text{ as } \mu \to 0
 \]
 we conclude \eqref{eq:limit_delta_g}. Therefore, for $\mu$ being sufficiently close to $0$ we have
 \[
 \tfrac{d}{dt}\|x\|_{\tilde \dn}\leq -\frac{\gamma 
 }{2\beta}\|x\|_{\tilde \dn}^{1+\mu}, \quad \forall x\neq \zero.
 \]
 The proof is complete.

 The above theorem proves that
 \begin{itemize}
   \item any linear (well-tuned) PID can be transformed ("upgraded") to a homogeneous PID allowing a faster regulation (finite-time or fixed-time stabilization at the set-point independently of the sign of $\mu$);
   \item theoretically, the transformation mentioned above can be done using any 
   $\dn$-homogeneous norm in $\R^2$.
 \end{itemize}
The potential usefulness of transforming a PID to hPID in practice will be studied for a robotic manipulator \cite{cervantes2001pid}.
\section{Application of hPID control to robotic manipulators}

\subsection{The selected mobile manipulator}

\begin{figure}[!ht]
    \centering
    \includegraphics[width=0.95\linewidth]{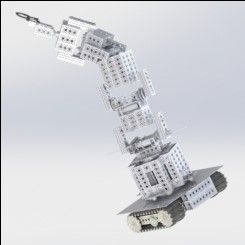}
    \caption{General view of the design of the mobile robotic manipulator. }
    \label{fig:robot arm}
\end{figure}

The structure of the robotic system for manipulating objects is shown in Figure \ref{fig:robot arm}. It consists of two principal structures. The first part is a robotic arm with six degrees of freedom and a gripper as its end effector, which can handle objects of up to 2kg. The second part corresponds to an autonomous vehicle that steers. 

\subsection{Mechanical configuration of the mobile manipulator}

The complete first degree of freedom of the robotic arm is shown in Figure \ref{fig:robot arm}. It has four ball wheels to reduce friction when the motor is activated. The addition of wheels is a modification of the original design, which also supports the weight that the robotic arm is carrying. The base of the platform also has a Hall effect sensor to find a start position and the angular velocity analysis. 

\subsection{Electrical instrumentation of the mobile manipulator}

\begin{figure}
    \centering
    \includegraphics[width=0.95\linewidth]{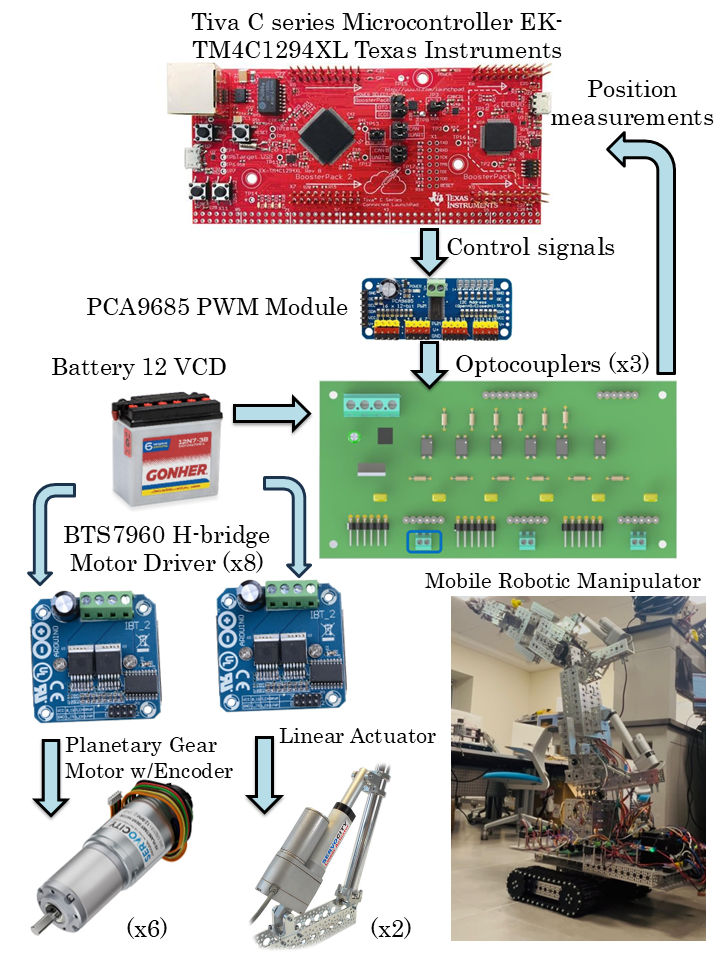}
    \caption{Simplified diagram of electrical instrumentation for actuating/measuring the activity of each joint in the mobile manipulator. }
    \label{fig:electrical inst}
\end{figure}

The instrumentation of the robotic arm is the same for the 6 DoF. Figure \ref{fig:electrical inst} shows a diagram of the connections for controlling and measuring the motors. It consists of five elements: the microcontroller, where the controller is implemented; the optocoupler to ensure safe electrical operation; the H-bridge circuit, which controls the DC motor direction; the motor encoder; and the Hall effect sensor. The microcontroller measures and activates the set of eight motors.

\subsection{Implementation of hPID on the mobile manipulator}

This subsection presents models that describe the position of the end-effector in the mobile manipulator with respect to the inertial frame and the motion dynamics of the geometric coordinates of the robotic mobile manipulator. 

The first model uses the variable $r_{RS} \in \mathcal{R}_{R} \oplus \mathcal{R}_{A} \subset \mathbb{R}^{6}$ with components $r_{RS}=\left[\begin{array}{cccccc}
x_{RS},&y_{RS},&z_{RS},&\theta_{RS},&\varphi_{RS},&\psi_{RS} \end{array} \right]^{\top}$ which is describing the relative position of the arm´s end-effector with respect to the inertial frame for the robotic manipulator.  

The functional relationship between the location vector of the mobile robot $r_{R}$ and the location vector of the end-effector $r_{a}$ with $r_{RS}$ can be obtained using a coordinate transformation given by:
 \begin{equation}
     r_{RS}(q_{R},q_{a},t)=Q_{z}(\theta_{a})r_{R}(q_{R},t) + r_{a}(q_{a},t),\label{RS position}
 \end{equation}
\noindent where $Q_{z} = diag\left\{ Rot_{z,\theta_{a}}, Rot_{z,\theta_{a}} \right\}$. Here $Rot_{z,\theta_{a}} $ is the canonical rotation matrix in the z-axis with respect to the inertial frame \cite{spong2020robot}.

The dynamics of $ r_{RS}(q_{R},q_{a})$ is governed by:

%
\begin{equation}
    \begin{array}{c}
         \dfrac{d^{2}}{dt^{2}} {r}_{RS}(q_{R},q_{a},t)=\dfrac{d^{2}}{dt^{2}}{Q}_{z}(\theta_{a})  r_{R}(q_{R},t) + \vspace{3mm} \\ 
         2\dfrac{d}{dt}{Q}_{z}(\theta_{a})  \dfrac{d}{dt}{r}_{R}(q_{R},t)  + \vspace{3mm} \\
          {Q}_{z}(\theta_{a}) \dfrac{d^{2}}{dt^{2}} {r}_{R}(q_{R},t)   +  \dfrac{d^{2}}{dt^{2}}{r}_{a}(q_{a},t).  
    \end{array}
    \label{VelocityDyamics}
\end{equation}

 Given the vector $r_{RS}$ and the dynamic expressions for the Euler-Lagrange for the robotic manipulator, the dynamic of the RS is:
 \begin{equation}
    \begin{array}{c}
         \dfrac{d^{2}}{dt^{2}}{r}_{RS}(q_{R},q_{a},t)= A_{RS} \left( \dfrac{d}{dt}{r}_{RS}, {r}_{RS} \right) + \\
         B_{RS} \left( q_{R},q_{a},t \right) u_{RS}(t)+\mathbf{\xi }_{RS}(t).  
    \end{array}
    \label{eq:DynMob}
 \end{equation}

Here 
\begin{equation*}
 \begin{array}{c}
  A_{RS} \left( \dfrac{d}{dt}{r}_{RS}, {r}_{RS} \right) = \dfrac{d^{2}}{dt^{2}} {Q}_{z}(\theta_{a})  r_{R}(q_{R},t) + \vspace{3mm}  \\
  2\dfrac{d}{dt}{Q}_{z}(\theta_{a})  \dfrac{d}{dt}{r}_{R}(q_{R},t) + \dfrac{d}{dt}{Q}_{a}(\theta_{a})\dfrac{d}{dt}{q}_{a}(t)+ \vspace{3mm} \\
  {Q}_{z}(\theta_{a}) M_{R} ^{-1}\left( q_{R} \right)A_{R} \left( \dfrac{d}{dt}{q}_{R},q_{R} \right) + \vspace{3mm} \\
      Q_{a}(r_{a})\tilde{M}_{a}^{-1}\tilde{A}_{a}\dfrac{d}{dt}{q}_{a}(t) \vspace{3mm} \\
     B_{RS} \left( q_{R},q_{a},t \right) = \\
        \left[ 
            \begin{array}{cc}
                {Q}_{z}(\theta_{a})M_{R}^{-1} \left( q_{R} \right)B_{R} \left(q_{R} \right)  &
                Q_{a}(r_{a})\tilde{M}_{a}^{-1}\tilde{B}_{a}(r_{a})             
            \end{array}
        \right]. \vspace{3mm}\\
     \mathbf{\xi }_{RS} ={Q}_{z}(\theta_{a})M_{R}^{-1} \left( q_{R} \right)\xi_{R}(t)+Q_{a}(\theta_{a})\tilde{M}_{a}^{-1}\mathbf{\tilde{\xi}}_{a}(t) 
 \end{array}   
\end{equation*}

Here $q_R \in \mathbb{R}^2$ and $q_a \in \mathbb{R}^2$ are the vectors of generalized coordinates for the vehicle and the arm. Matrices $\tilde{M}_{R}$ and $\tilde{M}_{a}$ are the inertia matrices for the vehicle and the manipulator. 

Matrices $A_{R}$ and $\tilde{A}_{a}$ are the drift terms of the vehicle and the arm, respectively. Complementary, $B_{R}$ and $\tilde{B}_{a}$ are the matrices that relate the individual control actions for the vehicle and arms, respectively. The extended control action $u_{RS} \in \mathbb{R}^{8}$ is $u_{RS} = \left[ u^{\top}_{R}, u^{\top}_{a} \right]^{\top}$. 

Considering the definition of an extended vector of generalized  coordinates $q_{RS}$:
$
     q_{RS}=\left[\begin{matrix}
         q_{R}&q_{a}
     \end{matrix}
     \right]^{\top} $, one may notice that there is a relation of the RS position $r_{RS}$ with the extended vector $q_{RS}$ considering the expression \eqref{RS position}, subject to the expressions direct kinematics expression of the robot arm and the ground vehicle.
 %
%

 Now,  considering the complete dynamic of the robotic arm on function of the generalized coordinates $q_{RS}$, and considering the dynamic equations of the joints of the robotic arm in equation \eqref{eq:DynMob}, is given by:
 \begin{equation}
 \begin{array}{c}
      \dfrac{d^{2}}{dt^{2}} {q}_{RS}(t)=M_{RS}^{-1} \left( {q}_{RS} \right) A_{RS}\left( \dfrac{d}{dt}{q}_{RS}, q_{RS} \right)  +  \\
      M_{RS}^{-1} \left( {q}_{RS} \right) \left[ \tilde{B}_{RS}u_{RS}+\mathbf{\xi}_{RS}\right]
 \end{array}     
     \label{Generalized coordinates} 
 \end{equation}

Define the extended vector $\zeta_{a} = {r}_{RS}(q_{R},q_{a},t)$ and its dynamics $ {\zeta}_{b} = \dfrac{d}{dt}{r}_{RS}$. According to the dynamics presented in \eqref{Generalized coordinates}, the dynamics of ${\zeta}_{a}$ is given by:
\begin{equation}
    \begin{array}{c}
         \dfrac{d}{dt}{\zeta}_{a}(t) \text{=} \zeta_{b}(t) \\
         \dfrac{d}{dt}{\zeta}_{b}(t) \text{=} A_{RS} \left( \zeta_{b}(t), \zeta_{a}(t) \right) + B_{RS} \left( \zeta_{a}(t) \right) u_{RS}(t)+\mathbf{\xi }_{RS}(t) 
    \end{array}
    \label{DynamicsRS}
\end{equation}

Given the RS dynamics, the variable $\zeta_a$ must follow a reference trajectory to approach the $j-th$ block and position the gripper to grab a target object. 

In view of the dynamics for the mobile manipulator \eqref{DynamicsRS}, lets propose a feasible set of references associated with the motion of the joints for the robotic manipulator when it must handle the j-th brick, represented as $\zeta^{*,(j)}_{a} : \mathbb{R}^{+} \rightarrow \mathcal{R}_{R} \oplus \mathcal{R}_{A} \subset \mathbb{R}^{6}$. The related states $\zeta^{*,(j)}_{a}$ should satisfy:
\begin{equation}
\begin{array}{c}
\dfrac{d}{dt}{\zeta}^{*,(j)}_{a}(t) = \zeta^{*,(j)}_{b}(t), \vspace{1mm} \\
\dfrac{d}{dt}{\zeta}^{*,(j)}_{b}(t) = h \left( \zeta^{*,(j)}_{a}(t), \zeta^{*,(j)}_{b} (t) \right).
\end{array}\label{sysref}
\end{equation}

Here $h :  \mathcal{R}_{R} \oplus \mathcal{R}_{A}  \times \mathcal{T} \mathcal{R}_{R} \oplus \mathcal{T}\mathcal{R}_{A}  \rightarrow \mathbb{R}^{6}$ is a class of smooth vector fields that defines the class of admissible reference trajectories related to the needed motion for the corresponding brick. 

The aforementioned leads us to formulate the error in path tracking $\delta^{(j)}_{a}$ defined as a function of $\zeta_{a}$ and $\zeta^{*,(j)}_{a}$ as follows:
\begin{equation}
\delta^{(j)}_{a}=\zeta^{*,(j)}_{a} - \zeta^{(j)}_{a}
\end{equation}

Given the mathematical model of the RS developed using the Euler-Lagrange analytical mechanics theory in and the reference description presented in \eqref{sysref}, the dynamics of the tracking error can be described with the following ordinary differential equation:
\begin{equation}
\begin{array}{c}
\dfrac{d}{dt}{\delta}^{(j)}_{a}(t)=\delta^{(j)}_{b}(t), \vspace{2mm} \\
\dfrac{d}{dt}{\delta}^{(j)}_{b}(t)= h^{(j)} \left( \zeta^*_{a}(t), \zeta^*_{b} (t) \right)  -  A^{(j)}_{RS} \left( \dfrac{d}{dt}{r}_{RS}(t), {r}_{RS}(t) \right)  \vspace{2mm} \\
- \dsum_{k=1}^{8} 
 B^{(j,k)}_{RS}(t) u^{(k)}_{RS}(t)-\mathbf{\xi }^{(j)}_{RS}(t) . 
\end{array}\label{sys1}
\end{equation}
Here, the variable $\delta^{j}_{a}$ refers to the tracking error for the j-th generalized coordinated, while $\delta^{j}_{b}$ corresponds to the time derivative of the mentioned tracking error described in the suggested generalized coordinates. The mechanical structure for the robotic arm implies that $\xi^{j}_{RS} (t) = \xi^{j}_{RS} \left( \zeta_{b}(t),\zeta_{a}(t),t\right)$ satisfies the next inequality (which implies the corresponding bounds for perturbations and uncertainties):
\begin{equation}
    \begin{array}{c}
         \left\vert \xi^{j}_{RS} \left( t\right) \right\vert \leq \xi_{a}, \enskip \forall t \geq 0,  \qquad
         \xi_{a} > 0   
    \end{array}
    \label{UpperboundXi}
\end{equation}

Considering that $B^{(j,j)}_{RS}$ is invertible for all $\zeta_{a},$ then select 
\begin{equation}
 \begin{array}{c}
      u^{(k)}_{RS} = - \left( B^{(j,j)}_{RS} \right)^{-1} \left( \dsum_{k=1, k\neq j}^{8}  B^{(j,k)}_{RS} u^{(k)}_{RS} \right) +  \\
       \left( B^{(j,j)}_{RS} \right)^{-1} \left( h^{(j)} \left( \zeta^*_{a}, \zeta^*_{b}  \right) - A^{(j)}_{RS} \left( \dfrac{d}{dt}{r}_{RS}, {r}_{RS} \right) \right) - \\
      K^{j}_{p} {\delta}^{(j)}_{a} - K^{j}_{d} {\delta}^{(j)}_{b} - K^{j}_{i} \int^{t}_{\tau=0} {\delta}^{(j)}_{a} d \tau
  \end{array}   
\end{equation}
 The substation of this control action in \eqref{sys1} leads to 
\begin{equation}
\begin{array}{c}
\dfrac{d}{dt}{\delta}^{(j)}_{a}(t)=\delta^{(j)}_{b}(t), \vspace{2mm} \\
\dfrac{d}{dt}{\delta}^{(j)}_{b}(t)=K^{j}_{p} {\delta}^{(j)}_{a}(t) + K^{j}_{d} {\delta}^{(j)} (t) +  \\
 K^{j}_{i} \dint^{t}_{\tau=0} {\delta}^{(j)}_{a} d \tau - \mathbf{\xi }^{(j)}_{RS}(t) . 
\end{array}\label{sys1}
\end{equation}
 
This structure corresponds to the formulation of PID structure with output ${\delta}^{(j)}_{a}$. A collection of eight controllers is implemented in a decentralized manner for the selected mobile manipulator.  

\section{Experimental assessment of HPID on the mobile manipulator }

In the experimental study, we consider the following homogeneous norm $\|\xi\|_{\dn} = \zeta^{-1}_{1,max} \vert \zeta_{1} \vert^{1/(1 - \mu)} + \gamma \vert \zeta_2 \vert$ where $ \zeta = \left[ \zeta_{1}, \zeta_{2}\right]^{\top}$, and $\zeta_{1,max}$ is the maximum value of $\zeta_1$. In the case of the distributed hPID control application on the mobile manipulator, $\zeta_{1}$ is the tracking error and $\zeta_{2}$ is the derivative of the tracking error for each joint. 

To assess control system performance, this work considers the analysis of three key indices, each capturing a distinct aspect of system behavior. The Integral Variation of the Control Signal (IVC) quantifies the abruptness and variability of the control input, helping identify excessive fluctuations that may compromise actuator integrity or destabilize the system. The Integral of the Absolute Value of the Control Signal (IAVC) reflects the total control effort exerted, offering a measure of energy consumption and actuation intensity. Meanwhile, the Integral of Time-weighted Absolute Error (ITAE) emphasizes the duration and magnitude of tracking errors, particularly during the settling phase, and helps reduce the impact of large initial deviations. Together, these indices provide a comprehensive evaluation of control efficiency, smoothness, and accuracy.
The following expressions define these indices used to perform the comparison of the selected PID controllers: 

\begin{equation*}\label{ise}
	\begin{array}{ccccc}
\displaystyle	IVC =  \int_{0}^{9} \left| \dfrac{du}{d t}\right| d t, \quad  \displaystyle	 IAVC =  \int_{0}^{9} \left| u\right| d t, \quad \vspace{4mm} \\[0.5cm]
\displaystyle	 ITAE= \int_{0}^{9}  t \left|  \epsilon\right| d t   
	\end{array} 
\end{equation*}

The selection of these indices is justified, considering the need to implement a fair comparison with traditional PID, which is effective in reducing IVC, IAVC, and ITAE. 

\begin{figure}[b!]
	\begin{center}
		\includegraphics[trim={2.0cm 2.5cm 1.5cm 2.5cm},clip,scale=0.36]{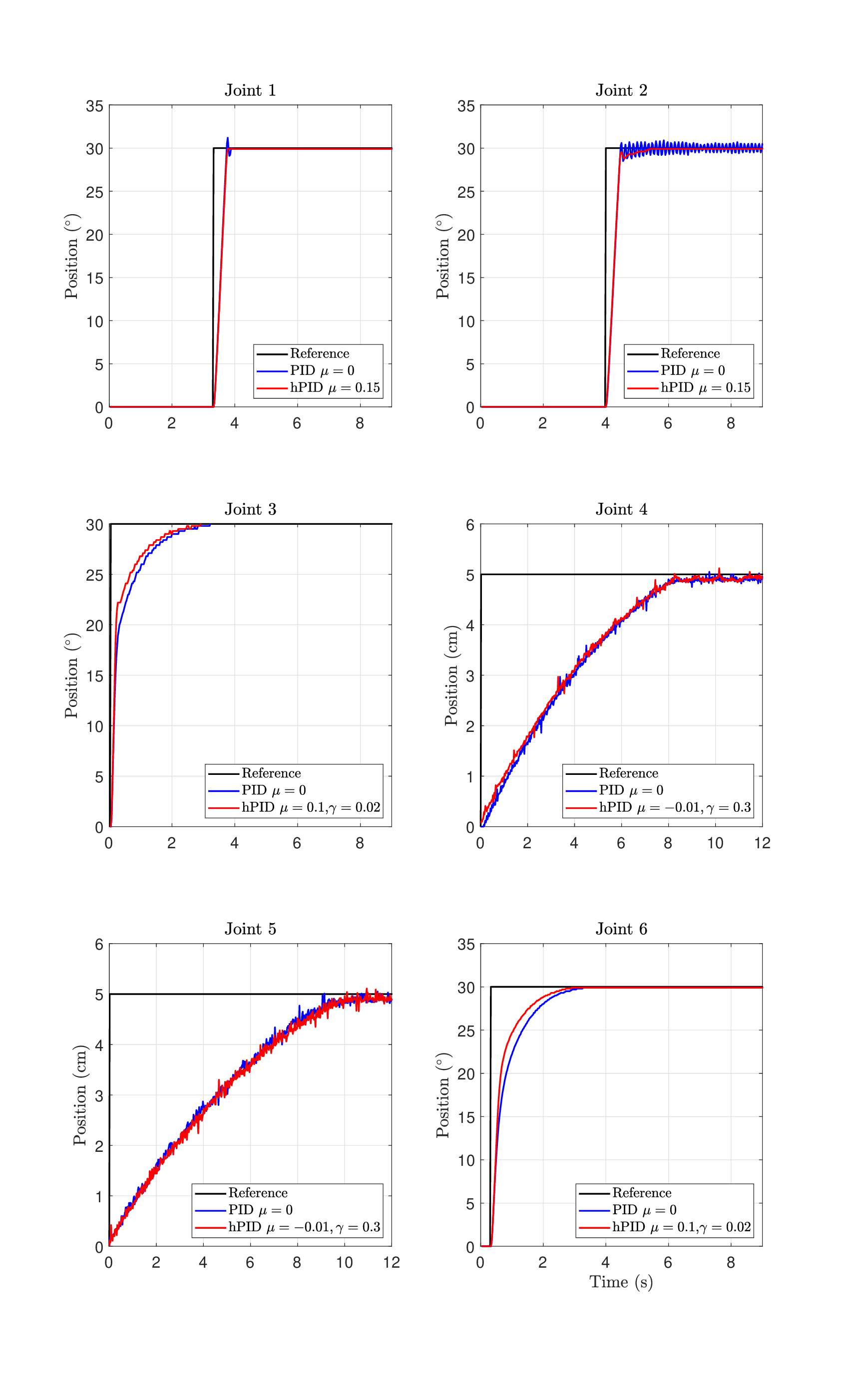}
		\caption{Comparison of the temporal evolution of trajectories for all joints in the arm section of the mobile manipulator produced with the implementation of the hPID (red lines) and PID (blue lines).}
		\label{FIG01}
	\end{center}
\end{figure}
\begin{figure}[b!]
	\begin{center}
		\includegraphics[trim={1.8cm 2.5cm 1.5cm 2.5cm},clip,scale=0.36]{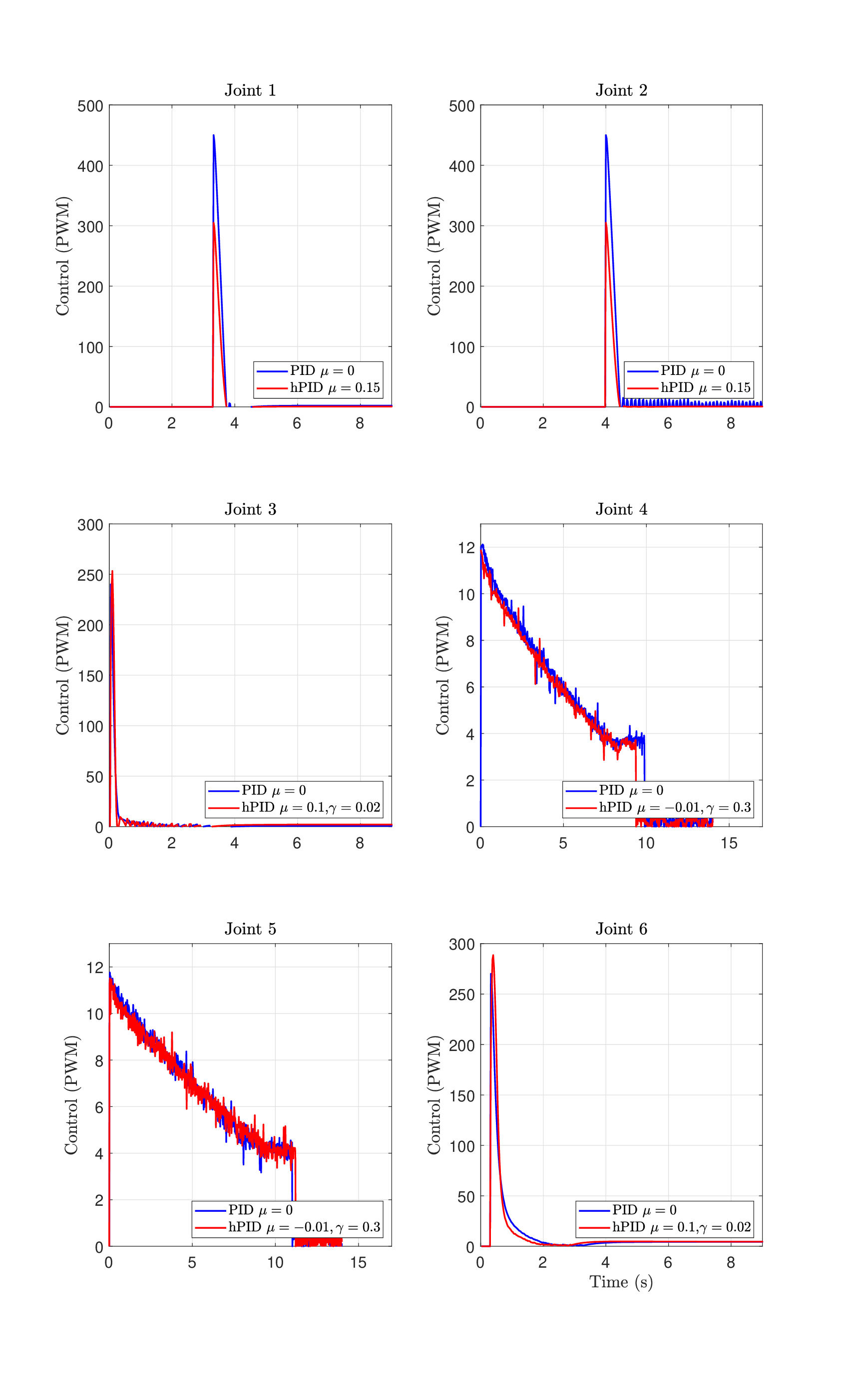}
		\caption{Comparison of the temporal evolution of control functions for all joints in the arm section of the mobile manipulator produced with the implementation of the hPID (red lines) and PID (blue lines).}
		\label{FIG02}
	\end{center}
\end{figure}

\begin{figure}[b!]
	\begin{center}
		\includegraphics[trim={2.0cm 2.5cm 1.5cm 2.5cm},clip,scale=0.37]{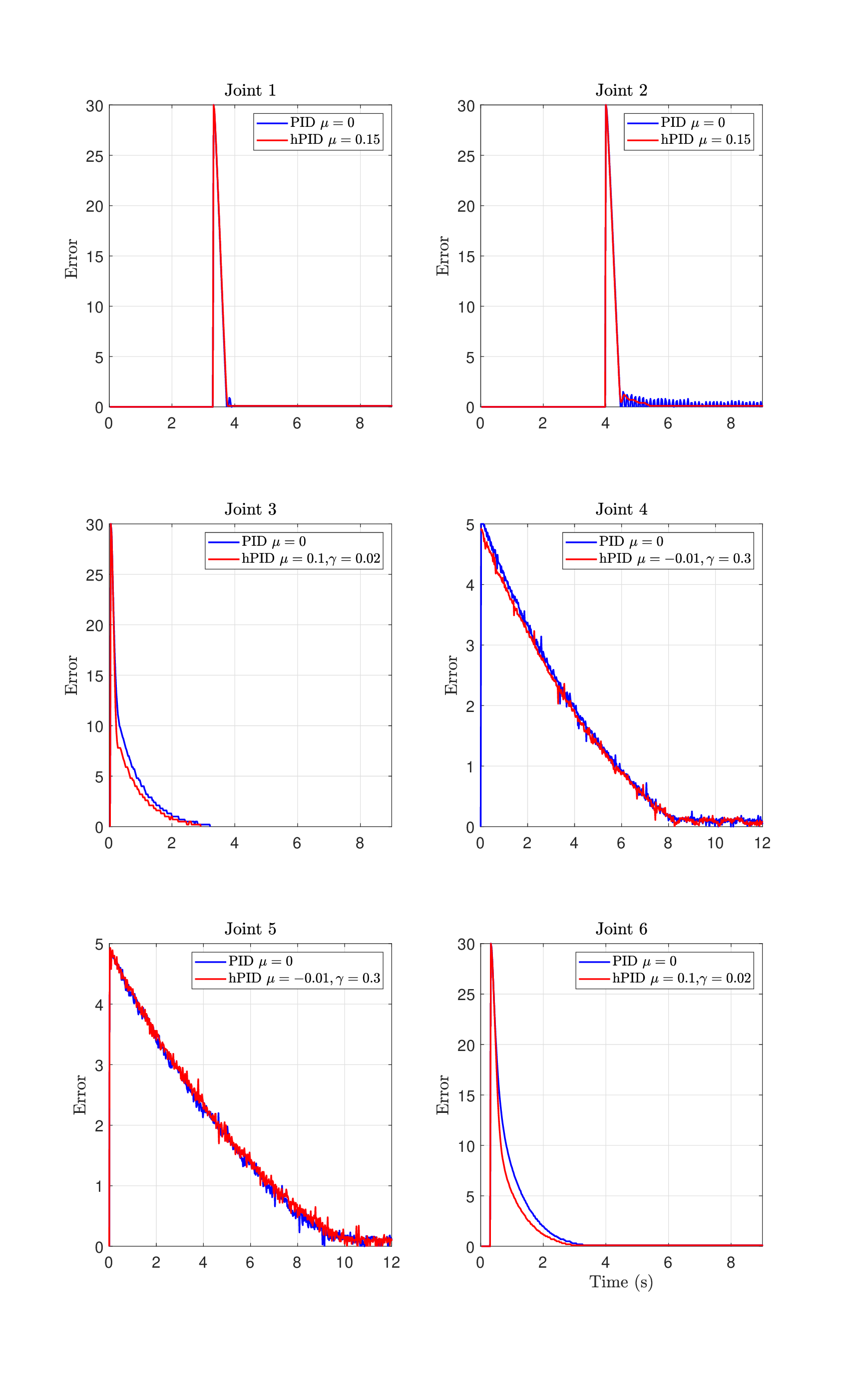}
		\caption{Comparison of the temporal evolution of trajectory tracking errors for all joints in the arm section of the mobile manipulator produced with the implementation of the hPID (red lines) and PID (blue lines).}
		\label{FIG03}
	\end{center}
\end{figure}

\begin{figure}[tb!]
	\begin{center}
		\includegraphics[scale=0.2]{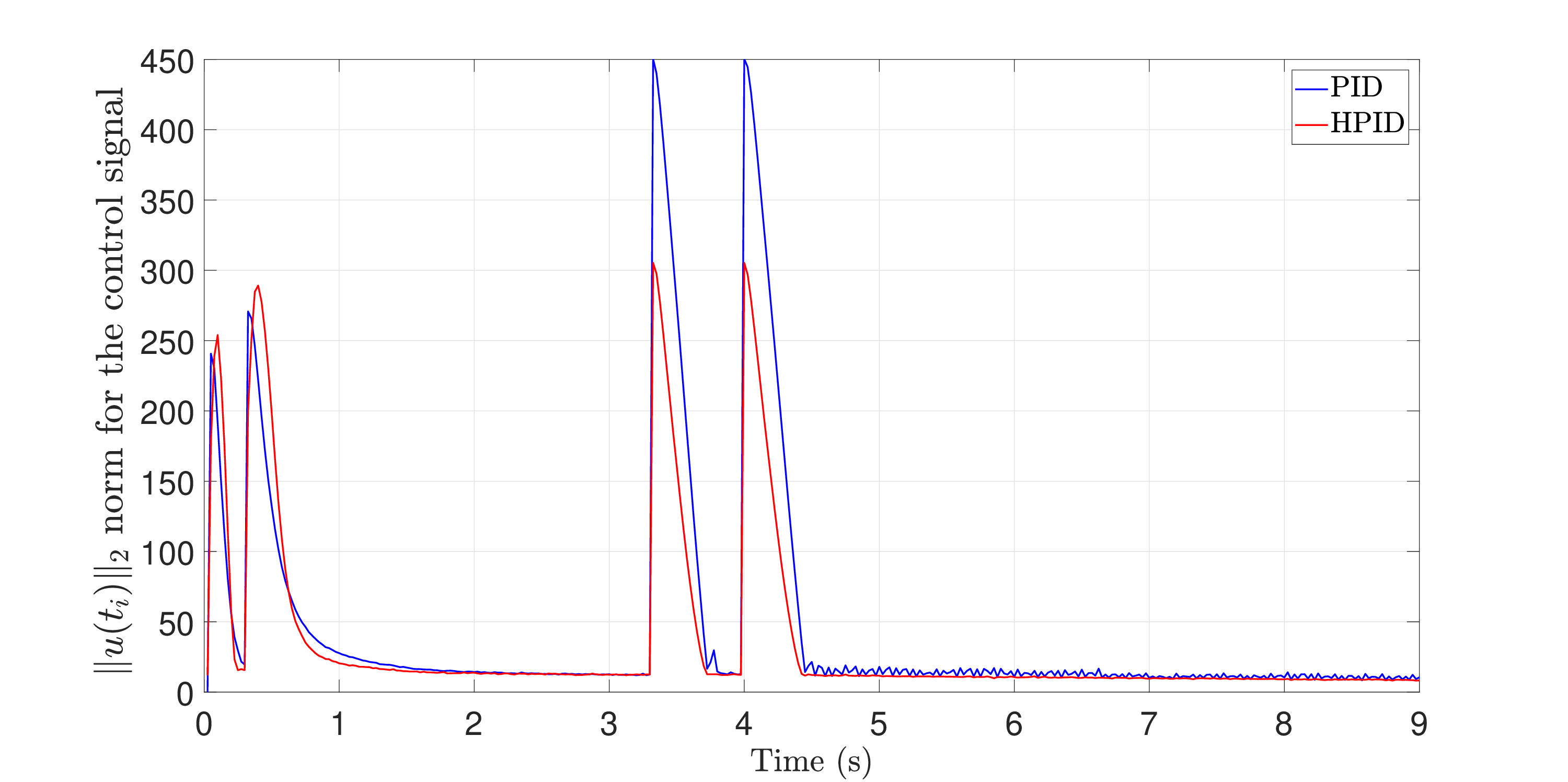}
		\caption{Comparison of the temporal evolution for $\|u(t_i) \|_2$ norm with the implementation of the hPID (red lines) and PID (blue lines).}
		\label{FIG03}
	\end{center}
\end{figure}

\begin{figure}[tb!]
	\begin{center}
		\includegraphics[scale=0.2]{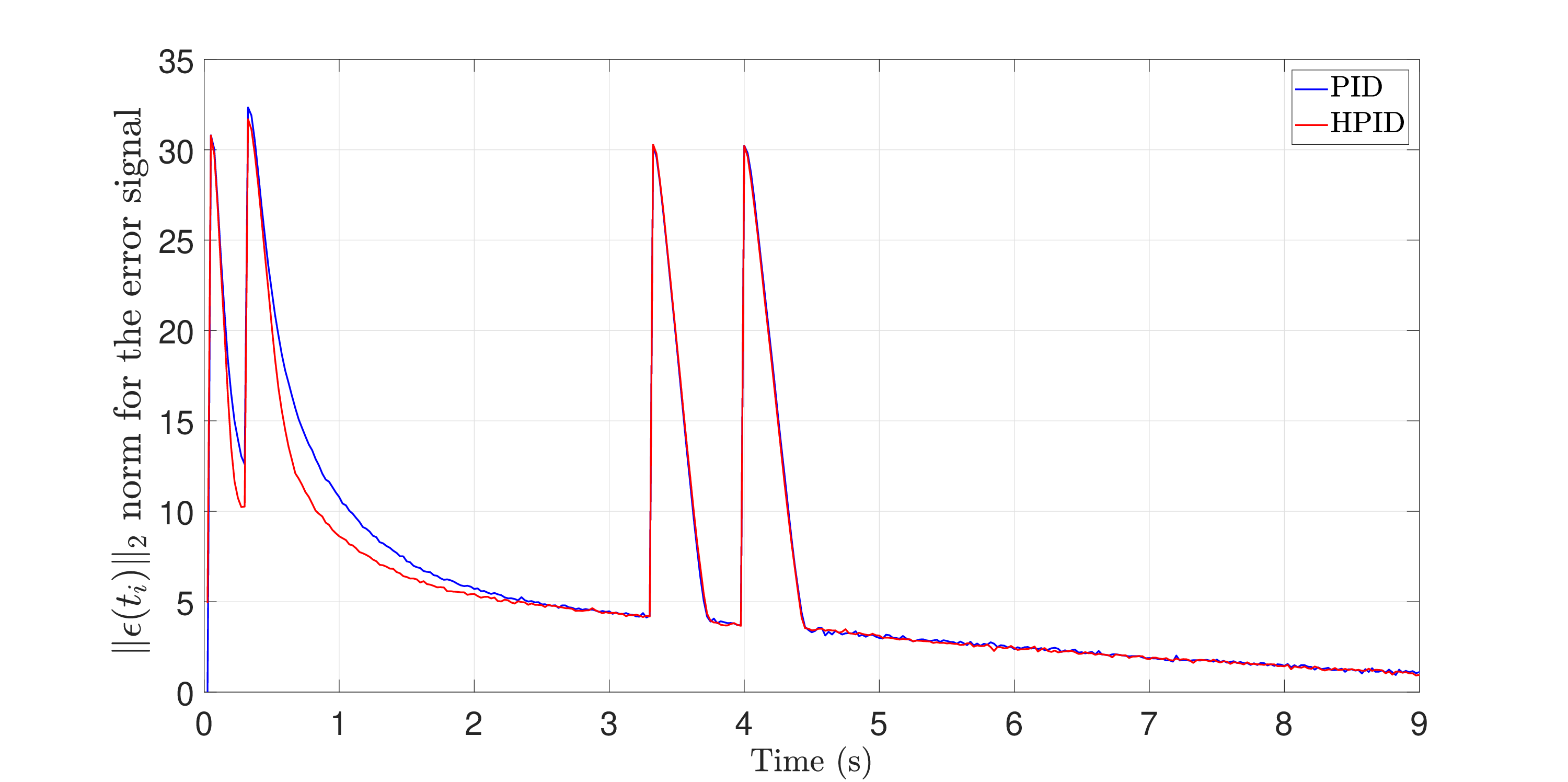}
		\caption{Comparison of the temporal evolution for $ \| \epsilon(t_i) \|_2$ norm with the implementation of the hPID (red lines) and PID (blue lines).}
		\label{FIG03}
	\end{center}
\end{figure}

   \begin{table}[b!]
	\centering
	\caption{Performance index comparison between PID and HPID controllers.}
	\begin{tabular}{ |c|c|c|c|c|c|c| }
		\hline
		Joint & \multicolumn{2}{c|}{$IVC$} & \multicolumn{2}{c|}{$IAVC$} & \multicolumn{2}{c|}{$ITAE$} \\ \hline
		        & PID       & HPID      & PID      & HPID      & PID      & HPID      \\ \hline
		Joint 1 & 982       & 615       &114       & 65        & 2.776    & 2.806         \\ \hline
		Joint 2 & 2390      & 618       & 142      & 68        & 4.385    & 3.704   \\ \hline
		Joint 3 & 836      & 853      & 43      & 50        & 1.198    & 0.987   \\ \hline
		Joint 4 &  111      &  56      & 67       & 65        & 6.589    & 6.345   \\ \hline
		Joint 5 &  109      &   133     & 73       & 72        & 7.925    & 8.150    \\ \hline
		Joint 6 & 552       & 591       & 102       & 108        & 1.802    & 1.3733   \\ \hline
	\end{tabular}
	\label{TABLE_PID_HPID}
\end{table}

This section also presents a comparative evaluation of the PID and hPID controllers applied to a six-joint robotic system, utilizing three performance indices: IVC, IAVC, and ITAE. These indices assess control smoothness, energy expenditure, and error persistence over time, respectively. The results are summarized in Table 1, and the system's responses are illustrated in Figures 3 to 5.

Table \ref{TABLE_PID_HPID} presents a comprehensive joint-by-joint comparison of control indices for both PID and hPID controllers, highlighting their performance differences in various metrics. The hPID controller exhibits substantial improvements in control smoothness and energy efficiency across all evaluated joints. For instance, in Joint 2, IVC decreases sharply from 2390 for the PID controller to just 618 with the hPID configuration, while the IAVC also reflects a significant drop from 142 to 68. These reductions indicate a noteworthy decrease in control signal variability and energy effort required to achieve desired joint movements.

The tracking trajectory improvements are visually corroborated in Figure 4, where the control signals generated by the hPID controller display markedly fewer abrupt transitions and a lower amplitude in comparison to their PID counterparts. The smoothness of the hPID signals suggests a more stable and less aggressive control strategy, which is likely to prolong the lifespan of the actuators by reducing mechanical stress during operation.

When assessing tracking accuracy, the performance of the hPID controller is further validated by its lower Integral of Time-weighted Absolute Error (ITAE) values across four out of six joints, specifically in Joints 2, 3, 4, and 6. This reduction in ITAE implies that the hPID controller enables faster and more precise convergence to the reference trajectories during operation. The distinct advantages of this controller are clearly illustrated in Figure 5, where the error signals produced with the implementation of the hPID controller exhibit a more rapid decay towards the origin, maintaining consistently lower magnitudes throughout the entire simulation window. Although Joints 1 and 5 display slightly elevated ITAE values under HPID control, the discrepancies are minimal and do not adversely impact the overall performance.

Additionally, the trajectories depicted in Figure 3 provide further evidence supporting these findings. Under the application of the hPID control, the joint responses show remarkable smoothness and a more accurate alignment with the reference signals, particularly during the transient phase of motion. This alignment indicates an enhanced dynamic behavior, characterized by reduced overshoot and oscillation, which is critical in applications demanding precision, such as the case for adjusting the position of mobile manipulators.

The implementation of the hPID controller not only leads to refined actuation strategies but also emphasizes an effective balance between smooth operation and precise trajectory tracking. Its ability to diminish the norm of the tracking error amplitude while simultaneously lowering energy consumption without compromising accuracy makes it a highly attractive alternative to traditional PID control. This is especially relevant in fields where actuator health and energy efficiency are of paramount importance, such as robotics, automotive systems, and aerospace applications. Overall, these findings underscore the potential of hPID control strategies to enhance system performance while ensuring reliability and efficiency in complex control applications. This is a remarkable characteristic of the proposed controller, considering that a better reference tracking is attained without using more control energy as shown by the variation of the selected indeces.

Figure 6 presents the plot of the  $\| \mathbf{u}(t_i) \|_{L^2} $ norm of the control signals applied to the six joints, computed over the interval \( t \in [0, 9] \). This norm serves as a compact measure of the overall control energy distributed across all actuators. The results indicate that the hPID controller requires less cumulative control effort, with a norm value of $\|u\|_{L^2(0,9)} = 329.459$, compared to $\|u\|_{L^2(0,9)} = 423.933$ for the PID. The norms are defined as:

\begin{equation}
 \|u(t_i) \|_2   = \left( \sum_{j=1}^{6} u_j^2(t_i) \right)^{1/2} 
\end{equation}

\begin{equation}
  \|u\|_{L^2(0,9)} = \left( \int_{0}^{9} \sum_{i=1}^{6} u_i^2(t) \, dt \right)^{1/2}  
\end{equation}

where \( u_i(t) \) represents the control signal for joint \( i \).

Conversely, Figure 7 shows the $  \| \mathbf{\epsilon}(t_i) \|_{L^2} $ norm of the position error signals across all joints. In this case, the PID controller achieves a slightly lower error norm, with $\| \epsilon  \|_{L^2(0,9)} = 52.2818$  , while the hPID yields  $\| \epsilon  \|_{L^2(0,9)} = 55.0789 $. This suggests that although hPID improves control smoothness and energy efficiency, it may introduce marginally higher residual error in certain joints during the transient period of the hPID evolution. The error norms are defined similarly as follows:

\begin{equation}
 \| \epsilon(t_i) \|_2  = \left( \sum_{j=1}^{6} u_j^2(t_i) \right)^{1/2} 
\end{equation}

\begin{equation}
  \| \epsilon  \|_{L^2(0,9)}  = \left( \int_{0}^{9} \sum_{i=1}^{6} \epsilon_i^2(t) \, dt \right)^{1/2} 
\end{equation}
where $ \epsilon_i(t) $ represents the position error for joint \( i \). These results reinforce the trade-off between control effort and tracking precision, highlighting the importance of selecting a controller based on application-specific priorities.


Using the information in Figure 4, corresponding to the position trajectories for each joint under both the PID and hPID control schemes, it is noticeable that the PID controller shows significant overshoot and oscillations, particularly in joints that have a higher dynamic demand that must handle sections in the manipulator with larger inertia. In contrast, the proposed hPID controller effectively reduces the amplitude of these oscillations, resulting in smoother and more stable trajectories. This qualitative improvement is supported by quantitative indices. The selected hPID consistently achieves lower values for both the IVC and the IAVC, indicating reduced control aggressiveness and improved energy efficiency. Furthermore, the $\| \epsilon  \|_{L^2(0,9)}$ norms of position error signals are quite close, with $\| \epsilon  \|_{L^2(0,9)}= 59.214$ for PID and $\| \epsilon  \|_{L^2(0,9)} = 55.865$ in the case of hPID. Collectively, these results indicate that the hPID controller maintains comparable error performance while significantly enhancing control smoothness and reducing actuation effort.

\section{Conclusion}

This study presents a comprehensive evaluation of PID and hPID controllers applied to a six-joint robotic system, using both error performance indices and energy-based metrics. The results demonstrate that the hPID controller consistently improves control smoothness and reduces energy consumption, as evidenced by lower IVC and IAVC values across most joints. While the PID controller occasionally achieves slightly lower tracking error as reflected in the ITAE and $\| \epsilon  \|_{L^2(0,9)}$ error norms, the differences are marginal and do not outweigh the benefits of reduced control effort.

Visual analysis of the position trajectories (Figure 3), control signals (Figure 4), and error profiles (Figure 5) further supports these findings. The hPID controller effectively attenuates overshoot and oscillations, resulting in more stable and actuator-friendly behavior. Additionally, the $\|u\|_{L^2(0,9)}$ norm of the control signal (Figure 6) confirms a significant reduction in overall actuation energy with hPID, while the error norm (Figure 7) remains similar to that of PID.

Overall, the hPID controller offers a favorable trade-off between tracking accuracy and control efficiency, making it a robust alternative for applications where actuator longevity, energy constraints, or mechanical stability are critical. These insights can inform future controller design and tuning strategies, particularly in multi-joint robotic systems with heterogeneous dynamics.


%



\section*{Acknowledgment}

This research was financially supported by the Tecnologico de Monterrey Challenge-Based Research Program project ID IJXT070-23EG60002 and by the French National Research Agency, Project ANR-24-CE48-2771 SLIMDISC.

\section*{{Competing interests}}
 The authors declare no relevant financial or non-financial interests.\\

\section*{{Author contributions}}
All authors contributed to the study conception, design, material preparation, data collection, and analysis.

\section*{Data availability} 
The data sets generated during and/or analyzed during the current study are available from the corresponding author on reasonable request.

\ifCLASSOPTIONcaptionsoff
  \newpage
\fi

\bibliographystyle{plain}
\bibliography{bib_all.bib}

\end{document}